\documentclass[journal]{IEEEtran}
\usepackage[labelfont=bf,labelsep=period]{caption}
\usepackage{graphicx}
\usepackage{float}
\usepackage{geometry}
\usepackage{booktabs}
\usepackage{tabularx}
\usepackage{hyperref}
\usepackage{amsmath}
\usepackage{mathtools}
\usepackage{amsfonts}
\usepackage{amssymb}

\usepackage[noadjust]{cite}

\title{ECG-SleepNet: Deep Learning-Based Comprehensive Sleep Stage Classification Using ECG Signals}
\author{
  Poorya Aghaomidi\textsuperscript{1}, Ge Wang\textsuperscript{1} \\
  \vspace{0.5em}
  \textsuperscript{1}{\small Department of Biomedical Engineering and Center for Biotechnology and Interdisciplinary Studies, Rensselaer Polytechnic Institute, Troy, NY, USA}
}

\begin{document}

\maketitle

\begin{abstract}
Accurate sleep stage classification is essential for understanding sleep disorders and improving overall health. This study proposes a novel three-stage approach for sleep stage classification using ECG signals, offering a more accessible alternative to traditional methods that often rely on complex modalities like EEG. In Stages 1 and 2, we initialize the weights of two networks, which are then integrated in Stage 3 for comprehensive classification. In the first phase, we estimate key features using Feature Imitating Networks (FINs) to achieve higher accuracy and faster convergence. The second phase focuses on identifying the N1 sleep stage through the time-frequency representation of ECG signals. Finally, the third phase integrates models from the previous stages and employs a Kolmogorov-Arnold Network (KAN) to classify five distinct sleep stages. Additionally, data augmentation techniques, particularly SMOTE, are used in enhancing classification capabilities for underrepresented stages like N1. Our results demonstrate significant improvements in the classification performance, with an overall accuracy of 80.79\% an overall kappa of 0.73. The model achieves specific accuracies of 86.70\% for Wake, 60.36\% for N1, 83.89\% for N2, 84.85\% for N3, and 87.16\% for REM. This study emphasizes the importance of weight initialization and data augmentation in optimizing sleep stage classification with ECG signals.


\end{abstract}

\section{Introduction}
Sleep disorders are a persistent challenge throughout the human history. Despite significant advancements in medicine, the prevalence of these disorders remains high, and their impact on public health continues to be profound. This paradox may be partly explained by significant changes in modern human behaviors, particularly the widespread use of digital devices. Exposure to artificial blue light from screens disrupts our circadian rhythms, which are genetically programmed to be influenced by natural sunlight. Studies have shown that excessive exposure to blue light, especially during the evening and night, can suppress melatonin production, delay sleep onset, and disrupt the overall sleep cycle \cite{blue_light, room_light}. This may contribute to the increasing incidence of sleep disorders in the modern population, despite our advances in understanding and treating these conditions.

Sleep is not a simple binary state of being awake or asleep; rather, it is a complex process that occurs in stages. The American Academy of Sleep Medicine (AASM) and the Rechtschaffen and Kales (R\&K) criteria are the two primary standards for classifying sleep stages. The R\&K criteria, established in 1968, are the first standardized system for sleep stage classification. These criteria divide sleep into two main categories: Non-Rapid Eye Movement (NREM) sleep, which is further subdivided into stages N1, N2, N3, and N4, and Rapid Eye Movement (REM) sleep. NREM sleep stages progress from light to deep sleep, with stage N4 being the deepest and most restorative stage \cite{rechtschaffen1968}.

The AASM standards, introduced in 2007, refined the R\&K criteria. The AASM reduced the number of NREM stages from four (as in the original R\&K criteria) to three by merging stages 3 and 4 into a single stage (now called N3). This change was made based on evidence that these stages represent a similar level of sleep and are difficult to distinguish using traditional Polysomnography (PSG). The AASM guidelines also provides more detailed criteria for scoring arousals, respiratory events, and limb movements, which are essential for diagnosing sleep disorders \cite{berry2012}.

Traditionally, experts rely on PSG data to classify sleep stages. PSG is a comprehensive sleep study that records various physiological signals, including Electroencephalogram (EEG), Electrocardiogram (ECG), Electromyogram (EMG), and Electrooculogram (EOG). Analyzing these signals to determine sleep stages is a complex task, often requiring significant expertise. The transitions between stages are subtle and can be challenging to identify, leading to potential errors even for experienced clinicians.

The advent of machine learning and more recently deep learning has further advanced sleep stage classification. Lyu et al. integrated EEG and ECG data to create a multimodal feature combination for sleep stage classification. Using multiscale entropy and intrinsic mode function for EEGs and heart rate variability for ECGs, they achieved an accuracy of 84.3\% on the ISRUC-S3 dataset\cite{lyu2022}. 
Similarly, Tao et al. employed a Squeeze-Excite (SE) fusion mechanism to combine EEG and ECG signals for classification. Using a recurrent convolutional neural network (RCNN) and balanced sampling, their model achieved 77.6\% accuracy on the MIT-BIH dataset across six sleep stages\cite{tao2022}. Utomo et al. focused on imbalanced datasets, using Weighted Extreme Learning Machine (WELM) and Particle Swarm Optimization (PSO) for feature selection, achieving a mean accuracy of 78.78\% for REM, NREM, and Wake stage classification\cite{utomo2023}. Lastly, Lesmana et al. combined Extreme Learning Machine (ELM) with PSO, reporting testing accuracies of 82.1\%, 76.77\%, 71.52\%, and 62.66\% for 2, 3, 4, and 6-class classification models, respectively\cite{lesmana2023}.

Recent studies have explored the use of ECG signals for sleep stage classification as a less intrusive alternative to PSG, capitalizing on the ease of ECG data acquisition. Initial efforts primarily focused on binary classification. For instance, Bozkurt et al. achieved 87.12\% accuracy using a hybrid machine learning model with 10 features to distinguish between sleep and wake states\cite{bozkurt2021}. Erdenebayar et al. expanded this model to a three-class classification (awake, light, and deep sleep) using a GRU-based deep learning model, reporting an accuracy of 80.43\%\cite{erdenebayar2020}.

For multi-class classification, Matsumoto et al. employed a decision tree model, achieving 74\% accuracy\cite{matsumoto2020}. Surantha et al. adopted an RNN approach utilizing heart rate variability (HRV) features extracted from ECG signals, obtaining 77\% accuracy\cite{surantha2023}. Fikri et al. designed a random forest classifier, producing mean accuracies of 82.72\%, 77.63\%, 73.38\%, and 62.62\% for two, three, four, and six-class models respectively\cite{fikri2023}. Pini et al. trained a deep learning model across three datasets, yielding accuracies of 88\%, 82\%, and 73\% for two, three, and four-class models respectively, on the CinC dataset\cite{pini2022}. Lastly, Choi et al. developed a deep learning architecture integrating CNNs and transformers, achieving 60\% accuracy for five-stage classification\cite{choi2023}.

Despite these advances, several limitations persist. A large portion of research focused predominantly on binary sleep-wake classification, with less attention given to differentiating the five distinct sleep stages. Additionally, detecting the N1 sleep stage—an essential but challenging component of sleep stage analysis—remains inadequate for most approaches. Furthermore, many models report lower performance metrics in multi-class classification due to suboptimal architectures and training methods, limiting their clinical utility.

This study addresses the aforementioned gaps and offers the following key contributions:

\begin{itemize}
\item ECG-Only Approach for Real-World Usability: We developed an approach that uses only ECG signals, eliminating the need for complex modalities like EEG, making our research applicable to real-world scenarios.
\item Multi-Class Classification: Unlike most previous ECG-based studies, which target binary classification, our model distinguishes all five sleep stages, providing a more comprehensive analysis of sleep patterns.
\item Faster Convergence with Feature Imitating Network (FIN): Our FIN speeds up model convergence, addressing the slow convergence issue seen in other approaches.
\item High Accuracy in N1 Stage: By initializing our model with a binary classification task focused on detecting N1 sleep, we improve accuracy in this challenging stage.
\item State-of-the-Art Performance: Our optimized model architecture, which incorporates layers like Kolmogorov-Arnold Networks (KANs) and Liquid Neural Networks (LNN), combined with effective data oversampling methods, enables us to achieve superior performance metrics, significantly improving accuracy over previous studies.
\end{itemize}

    The remainder of this paper is structured as follows: The \hyperref[sec:methods]{Methods} section describes the dataset, model architecture, and parameters in our study. The \hyperref[sec:results]{Results} section presents the performance of our model. Finally, the \hyperref[sec:discussion]{Discussions} section discusses our results. Finally, the \hyperref[sec:conclusion]{Conclusion} summarizes our findings and mentions future work.

\section{Methods} 
\label{sec:methods}
\begin{figure*}
    \centering
    \includegraphics[width=\linewidth]{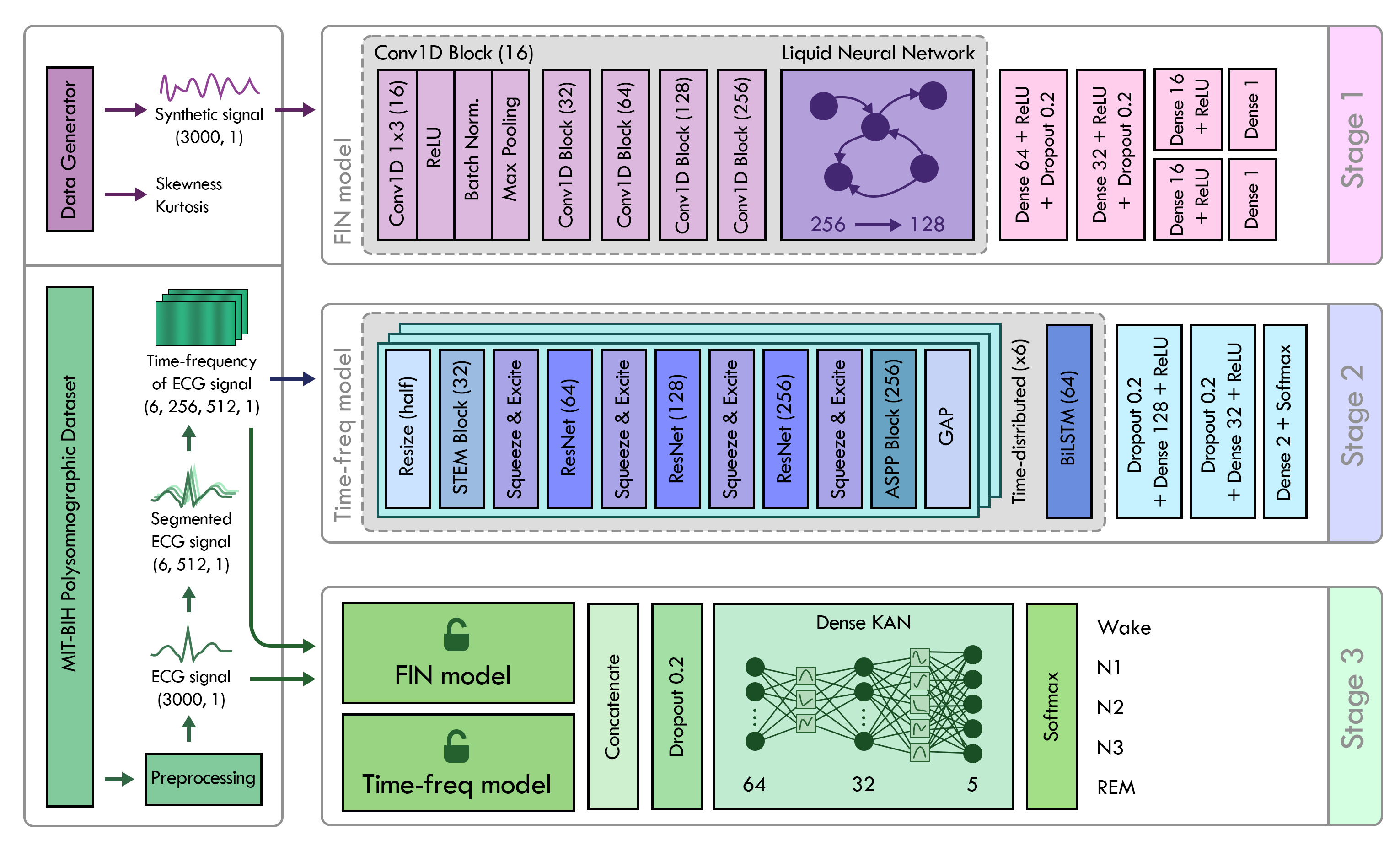}
    \caption{Overview of the three-stage sleep stage classification model. Stage 1 involves a FIN, utilizing an LNN architecture, trained on synthetic signals to estimate kurtosis and skewness. Stage 2 employs an architecture inspired by ResUNet++ to perform binary classification of N1 versus non-N1 stages using SST-based TFRs. In Stage 3, the pretrained FIN and the model used in Stage 2 are integrated using KANs for the final classification into Wake, N1, N2, N3, and REM stages.}
    \label{fig:model}
\end{figure*}

In this study, we propose a three-stage approach for sleep stage classification using ECG signals. The first phase involves estimating key statistical features, such as kurtosis and skewness. The second phase focuses on identifying the N1 sleep stage through the time-frequency representation of ECG signals. Finally, the third phase integrates the models trained in the previous stages and employs a Kolmogorov-Arnold Network to classify the five stages of sleep. An overview of this methodology is illustrated in Fig.~\ref{fig:model}.

\subsection{Dataset}
The dataset utilized in this study is sourced from the MIT-BIH Polysomnographic Database, which includes recordings of various physiological signals during sleep. These recordings were collected from subjects at Boston's Beth Israel Hospital Sleep Laboratory, where they were evaluated for chronic obstructive sleep apnea syndrome and the effects of Continuous Positive Airway Pressure (CPAP) therapy \cite{mitbih}  \cite{goldberger2000}.

The database comprises recordings from 18 subjects, each containing multiple channels of physiological data. For this study, we concentrated on the ECG signals, which were continuously recorded at a sampling rate of 250 Hz. The ECG data provides detailed insights into cardiac activities, which can be indicative of different sleep stages \cite{DeChazal2003}.

In addition to the ECG signals, the dataset includes reference annotations for sleep stages, labeled at 30-second intervals based on the R\&K criteria. The original sleep stage labels in the dataset include Wake (W), NREM stage 1 (N1), NREM stage 2 (N2), NREM stage 3 (N3), NREM stage 4 (N4), and Rapid Eye Movement (REM) sleep.

To prepare the ECG data for analysis, several steps were taken to ensure the signals be aligned with sleep stage labels. First, the original ECG signals, sampled at 250 Hz, were resampled to 100 Hz. This downsampling process reduced the number of data points in each 30-second segment from 7500 to 3000, effectively balancing computational efficiency with data fidelity. The critical features relevant to sleep stage classification were preserved, ensuring that the model had access to the necessary information while reducing computational overhead \cite{scipy2020}
To align with the more recent AASM criteria, which merges stages N3 and N4 into a single N3 stage, we combined these stages in our dataset. This adjustment ensures that our model's classification would adhere to contemporary sleep staging standards, facilitating comparability with other studies and clinical practices. The resulting sleep stage labels were then synchronized with the corresponding ECG segments, creating a cohesive dataset where each 30-second ECG segment was associated with an appropriate sleep stage label.

This standardized dataset, with AASM-compliant sleep stage labels, was used as the foundation for training and evaluating our deep learning models for sleep stage classification. The dataset consists of 3,115 wake signals, 1,815 N1 signals, 3,887 N2 signals, 669 N3 signals, and 700 REM signals. We split the dataset into 80\% for training, 10\% for validation, and 10\% for testing. The training portion was used to train our models in Stages 2 and 3.

\subsection{Stage 1}
FIN is a neural network designed to approximate one or more closed-form statistical features of a dataset. By initializing its weights, a FIN can achieve exceptional performance for various downstream signal processing tasks while requiring less data and fine-tuning compared to traditional networks\cite{DBLP:journals/corr/abs-2110-04831}. In this study, we trained our FIN to calculate kurtosis and skewness. These statistical features are crucial for analyzing ECG signals due to their sensitivity to outliers and abnormalities in the signal distribution. The purpose of the FIN pretraining was to initialize the weights of our model with values close to optimal rather than random initialization, providing a robust starting point for training on real ECG data.

\subsubsection{Data Synthesis}
To train the FIN, we generated synthetic signals composed of chirp signals, step signals, and random uniform noise. These types of signals were selected because, together, they span a wide range of kurtosis and skewness values. This diversity in the data allowed the FIN to learn to calculate these statistical measures under varying conditions. Each synthetic signal was 3,000 samples long, mirroring the length of real ECG data. The dataset included 20,000 signals for training, 2,000 for validation, and 2,000 for testing.

\subsubsection{Architecture}
The FIN model starts with five Convolutional Neural Network (CNN) blocks, each designed to progressively learn local and global features. Each block includes a 1D convolutional layer with a kernel size of 3 and filter numbers of 16, 32, 64, 128, and 256 in each respective block. Following the convolution, a ReLU activation function is applied, along with batch normalization, and a max pooling layer.

Following the CNN blocks, a Liquid Neural Network (LNN) was incorporated, which transformed the outputs from 256 to 128 units. LNNs are a type of neural network that differ from traditional recurrent neural networks (RNNs) by their ability to handle continuous, time-dependent data more effectively. Unlike RNNs that rely on static architectures with fixed time steps, LNNs use differential equations to model the dynamics of neurons over time, allowing them to adapt continuously to changes in input data. The "units" refer to the dimensionality of the feature space or the number of neurons in the LNN layer, where each unit represents a learned feature that contributes to the network's ability to process temporal patterns. By modeling neuron states as dynamic systems, LNNs can capture intricate time-varying relationships in the data, making them particularly effective for applications involving sequential or time-dependent inputs.

This design makes LNNs particularly suitable for analyzing ECG signals, which are naturally continuous and exhibit temporal dependencies. Given that our input length is 3,000 samples, the LNN’s dynamic nature is essential for maintaining the temporal context throughout the signal processing pipeline. This allows it to retain important features from earlier segments of the ECG signal while processing later parts, which is critical for tasks requiring an understanding of both short-term variations and longer-term trends in the data, such as for sleep stage classification.

Finally, fully connected layers responsible for the computation of kurtosis and skewness were added. Before branching into the two tasks, a common part consisting of a dense layer with 64 units followed by a ReLU activation function and a dropout layer with a rate of 0.2. This was followed by another dense layer with 32 units, also utilizing a ReLU activation function and a dropout layer with a rate of 0.2. For kurtosis, a dense layer with 16 units and a ReLU activation function was followed by another dense layer with 1 unit. The same structure was applied for skewness, enabling the network to compute these statistical features from the learned signal representations.

We trained the model using the Huber loss function with a batch size of 8 and the Adam optimizer at an initial learning rate of 0.001 over 20 epochs. To optimize training, we implemented ReduceLROnPlateau to dynamically decrease the learning rate when performance plateaued.

To evaluate the FIN after pretraining, we compared the true values of kurtosis and skewness from the synthetic signals with the predicted values generated by the FIN. Mean Squared Error (MSE) and Mean Absolute Error (MAE) were used as evaluation metrics to measure the model's accuracy in estimating these features. This step is crucial in confirming the FIN's capability to approximate these features before its integration into the sleep stage classification model.

\subsection{Stage 2}
The classification of the N1 sleep stage presents considerable challenges due to its transitional nature and overlap with adjacent stages. To address this, we introduced a dedicated stage in our model aimed at improving N1 detection. This stage involved a binary classification task that distinguishes N1 signals from non-N1 signals, helping to train a robust encoder tailored to capture the subtle features of N1.

\subsubsection{Preprocessing}
Given the imbalance in the dataset, we curated a balanced subset for the N1 classification task. The N1 class consisted of 1,815 signals, so we selected 454 signals from each of the remaining classes (N2, N3, REM, and Wake), creating a dataset with 1,815 N1 signals and 1,816 non-N1 signals.

Each signal, with a length of 3,000 samples (30 seconds), was windowed into six shorter signals with the length of 512. These windows were then transformed into time-frequency representations (TFRs) using the Synchrosqueezing Transform (SST)\cite{SST}. We used the "ssqueezepy" library for implementing the SST \cite{Muradeli2020ssqueezepy}. After applying SST, each 3,000-sample signal was represented as six TFRs, each with a size of 256x512, leading to an input shape of 6×256×512×1.

\subsubsection{Architecure}
To maintain continuity of the original 3,000-sample signals while processing the six time-frequency representation (TFR) windows, we employed a time-distributed approach, duplicating the encoder block six times to process each window in parallel. This strategy ensures that each TFR window is analyzed independently while still leveraging the information inherent in the original signal length.

For the encoder, we were inspired by the encoder used in the ResUNet++ architecture, known for its powerful feature extraction capabilities\cite{resunet}. The use of residual connections facilitates better gradient flow during training, addressing the vanishing gradient problem and enabling the model to learn both low-level and high-level features effectively. Additionally, using the stem block, Squeeze-and-Excitation (SE) blocks, and Atrous Spatial Pyramid Pooling (ASPP) block contribute to the model's ability to capture complex patterns in ECG signals, which are crucial for accurately identifying sleep stages. 

The encoder is structured to progressively extract features from the input image through a series of carefully designed blocks. It starts with a resizing layer to downsample the input dimensions, which prepares the data for subsequent processing. The first block is a stem block that applies convolutional layers to capture fundamental features with a stride of one, ensuring detailed spatial information is preserved. This is followed by three residual blocks, each designed to further refine the extracted features while doubling the number of filters at each stage to enhance the representational capacity. The residual connections in these blocks facilitate gradient computation promoting deep learning. At the end of each block, an SE block is included to adaptively recalibrate the feature responses, emphasizing informative features and suppressing less useful ones, thereby improving the model's overall performance.

\begin{figure}
    \centering
    \includegraphics[width=1\linewidth]{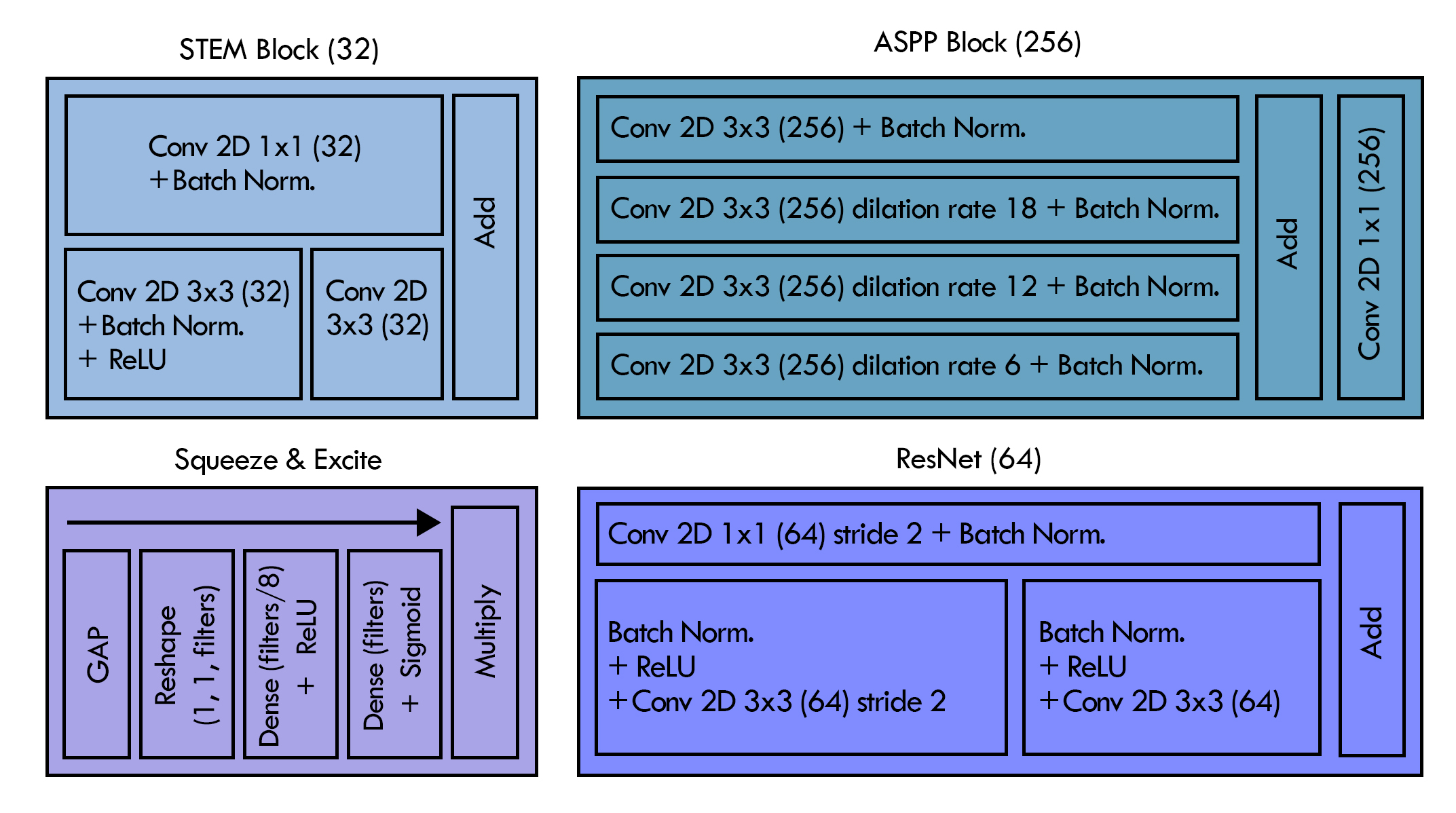}
    \caption{Detailed components of the ASPP, ResNet, SE, and STEM blocks used in the the encoder block.}
    \label{fig:n1}
\end{figure}

After the residual blocks, the encoder integrates an ASPP block, which employs multiple convolutional layers with varying dilation rates to capture multi-scale contextual information. This enhances the model's ability to recognize complex patterns in the input data. Finally, the architecture concludes with a Global Average Pooling layer, which consolidates the features extracted from the previous layers into a single feature vector for each window. This feature vector serves as a compact representation, facilitating subsequent classification tasks while maintaining the essential characteristics learned from the input. The specific order of these layers is shown in the second stage of Figure ~\ref{fig:model}, with detailed block components illustrated in Figure ~\ref{fig:n1}.

After processing the time-distributed encoder, the model utilizes a Bidirectional Long Short-Term Memory (LSTM) layer with 64 units to capture temporal dependencies in the sequential feature representations. This enables the model to leverage the past and future contexts from the data. The LSTM output is passed through a dense layer of 128 units with ReLU activation, followed by a dropout layer with a rate of 0.2 to reduce overfitting. A subsequent dense layer with 32 units further refines the learned features. The architecture concludes with a dense output layer consisting of 2 units and a softmax activation function, providing the final probabilities for classifying the presence or absence of the N1 sleep stage.

This stage of the model was trained using a batch size of 8, the categorical cross-entropy loss, and the SGD optimizer with a learning rate of 0.01. To optimize the learning process, we employed the ReduceLROnPlateau callback to adjust the learning rate dynamically according to the validation loss. The training was carried out for 100 epochs, ensuring the network effectively learned to distinguish N1 from the other sleep stages.

To evaluate the performance of this stage, we primarily used accuracy as a metric to assess how well the network differentiated N1 from the other sleep stages.

\subsection{Stage 3}
In the third stage of our model, we integrate the previously trained components to enhance the overall performance of sleep stage classification. This integration combines insights from the first two stages, where statistical features extraction and N1 classification were addressed independently. By employing the FIN, we leverage the powerful feature representations learned earlier, which facilitates faster convergence and improved performance across all sleep stages. Additionally, the model developed in the second stage, which focuses on the accurate identification of the N1 sleep stage, provides essential contextual information that further refines the classification process, leading to more reliable and accurate predictions of sleep stages overall.

\subsubsection{Preprocessing}
Sleep stages are inherently imbalanced, with certain stages such as N1, N3, and REM being underrepresented compared to others like Wake. This imbalance can negatively impact the performance of deep learning models, leading to biased predictions favoring the majority classes. To address this issue, we implemented three distinct augmentation techniques on the training set we split earlier to enhance the representation of the underrepresented classes and improve the robustness of our classification model:

\begin{itemize}
    \item SMOTE: Oversample all classes to match the size of the largest class (Wake, with 3,887 samples)\cite{smote}.
    \item ADASYN: Focus on oversampling the harder-to-classify minority classes by generating synthetic data in regions where the decision boundary is less distinct. In this case, ADASYN primarily oversampled the N3 class to 3,843 samples\cite{adasyn}.
    \item Custom Augmentation: Apply various augmentation techniques, including Time Warp, Random Quantize, Drift, and Reverse. This method yielded an equal representation across all classes with a total of 4,000 samples for each stage
\end{itemize}

As a result of these augmentation techniques, we created three distinct augmented datasets, alongside our original non-augmented dataset, totaling four datasets. We aim to test these four datasets to evaluate which augmentation approach contributes most effectively to improving classification performance across all sleep stages. It is important to note that the test and validation sets remain unchanged from \hyperref[sub sec:Dataset]{Section A}.

Since our final model in Stage 3 is a parallel integration of the pretrained models from the previous stages, and because they have two different inputs (1D signal and multi-time-frequency representations), we prepared the dataset with careful consideration of the unique requirements of each architecture. Two distinct input shapes were employed to ensure that each model received data tailored to its design:

\begin{itemize} 
    \item For the transferred model pretrained in Stage 1, we prepared 1×3,000 ECG signals and normalized them using min-max normalization to ensure consistent input values across the dataset. 
    \item For the transferred model pretrained in Stage 2, we performed the same preprocessing steps as outlined in that stage to obtain signals with a shape of 6×256×512×1. This involved windowing the original ECG signals and subsequently normalizing them using min-max normalization. 
\end{itemize}

By carefully preparing these two inputs from the ECG signals, we established a solid foundation for our hybrid model, enabling it to leverage the information from both Stages 1 and 2.

\subsubsection{Architecture}
In the final architecture, we streamlined the model by removing the fully connected layers from both the pretrained FIN and N1-detector models, leaving the remaining layers trainable. The outputs of these models were concatenated, followed by the application of a dropout layer with a rate of 0.2. The resulting concatenated feature vectors were then processed through a series of DenseKAN layers, which included a DenseKAN layer with 64 units, followed by another DenseKAN layer with 32 units, and a final DenseKAN layer with 5 units. The model concludes with a softmax activation function, responsible for classifying the input into five distinct sleep stages. Notably, the parameters of both the pretrained Stage 1 and 2 models, as well as the DenseKAN layers, remain fully trainable during the training process.

KANs represent an advancement in neural network design inspired by the Kolmogorov-Arnold representation theorem \cite{kolmogorov1957}. Instead of using fixed activation functions within nodes, KANs employ learnable activation functions on the edges or weights of the network\cite{liu2024kankolmogorovarnoldnetworks}. This flexibility allows KANs to better model complex patterns in the data, providing finer control over the learning process. This made them an ideal choice for the final layers of our model, as they optimize the classification system for ECG-derived sleep stages.

Our final stage was trained with a batch size of 8, using categorical cross-entropy as the loss function and the Adamax optimizer with an initial learning rate of 0.001, along with a learning rate reduction strategy and early stopping for 100 epochs.

We evaluated the performance of the model using standard classification metrics, including accuracy, precision, recall, and sensitivity, along with a confusion matrix and class-specific accuracy, ensuring a comprehensive assessment across all sleep stages.

The hardware and software configurations for model training and experimentation for all stages are specified as follows: the operating system Windows 11 with 32 GB of RAM, two RTX 3060 GPUs, each with 12 GB of memory, TensorFlow, version 2.9, and Python 3.9.

\begin{figure}[ht]
    \centering
    \includegraphics[width=\linewidth]{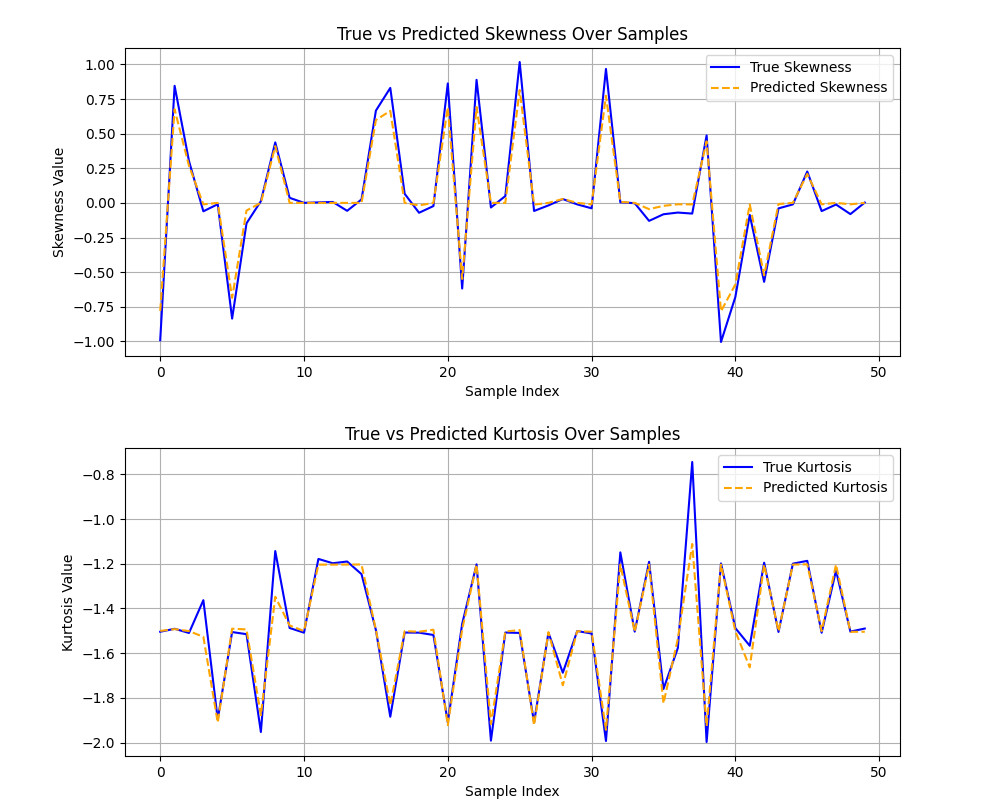}
    \caption{Predicted vs. True Values for Skewness (Top) and Kurtosis (Bottom).}
    \label{fig:kurtosis}
\end{figure}

\section{Results}
\label{sec:results}
In this section, we first present the performance of the FIN from Stage 1, followed by an evaluation of the N1 detection model from Stage 2. Lastly, we discuss the results of the final classification task, focusing on multi-class sleep stage classification. We assess the impact of different oversampling techniques, the effectiveness of incorporating KAN, LNN, and the role of weight initialization strategies employed in both Stages 2 and 3. Each of these elements contributed to optimizing the model's performance across all sleep stages.

\subsection{Stage 1 Results}
To evaluate the effectiveness of FIN, we compared the predicted and true values for skewness and kurtosis across synthetic signals. The model's performance in approximating these statistical features demonstrates its capacity to learn and generalize well on unseen data. As shown in Figures ~\ref{fig:kurtosis}, the predicted values closely match the true values.

The MSE and MAE for skewness and kurtosis during pretraining are as follows: for skewness, the MSE was 0.0066 and the MAE was 0.056, while for kurtosis, the MSE was 0.0059 and the MAE was 0.040.

\subsection{Stage 2 Results}
During the second stage, the model was trained specifically to discern between N1 sleep and the other classes, focusing on extracting relevant features for N1 identification. Despite the complexity of this task, the model achieved a balanced accuracy of 76.56\%. The precision, recall, and F1 scores for both classes are shown in Table ~\ref{tab:classification_report}.

\begin{table}[hb]
\centering
\caption{Classification Report for N1 Binary Classification.}
\begin{tabularx}{\linewidth}{X c c c c}
\toprule
\textbf{Class}           & \textbf{Precision} & \textbf{Recall} & \textbf{F1-Score} & \textbf{Support} \\ 
\midrule
\textbf{Class 0 (No N1)} & 0.83               & 0.63           & 0.72              & 150              \\ 
\textbf{Class 1 (N1)}    & 0.73               & 0.88           & 0.80              & 170              \\ 
\midrule
\textbf{Accuracy}        & -                  & -              & 0.77              & 320              \\ 
\textbf{Macro Avg}       & 0.78               & 0.76           & 0.76              & 320              \\ 
\textbf{Weighted Avg}    & 0.78               & 0.77           & 0.76              & 320              \\ 
\bottomrule
\end{tabularx}
\label{tab:classification_report}
\end{table}

\begin{table}[ht]
\centering
\caption{Performance Metrics for Different Augmentation Techniques using a model with full weight initialization (Top) and for Weight Initialization strategies using the SMOTE-oversampled dataset (Bottom).}
\begin{tabularx}{\linewidth}{X c c c c c c}
\toprule
\textbf{Dataset} & \textbf{Wake} & \textbf{N1} & \textbf{N2} & \textbf{N3} & \textbf{REM} & \textbf{Overall} \\
\midrule
Normal            & 81.33         & 51.64       & 83.19       & 50.51       & 62.39       & 73.29           \\
ADASYN            & 84.12         & 57.45       & 80.56       & 66.67       & 70.64       & 75.86           \\
Augmented         & 84.12         & 58.55       & 82.84       & 77.78       & 84.40       & 78.62           \\
SMOTE             & \textbf{86.70}         & \textbf{60.36}       & \textbf{83.89}       & \textbf{84.85}       & \textbf{87.16}       & \textbf{80.79}           \\
\midrule
\textbf{Init.}  & \textbf{Wake} & \textbf{N1} & \textbf{N2} & \textbf{N3} & \textbf{REM} & \textbf{Overall} \\
\midrule
No Init.          & 73.61         & 27.27       & 79.86       & 69.70       & 50.46       & 65.66 \\
No FIN Init.      & 84.76         & 59.09       & 83.74       & 77.64       & 85.49       & 78.14 \\
No N1 Init.       & 79.83         & 46.18       & 77.76       & 67.68       & \textbf{87.16}       & 72.70 \\
\bottomrule
\end{tabularx}
\label{tab:augmentation_results}
\end{table}

\subsection{Stage 3 Results: The Effect of Data Augmentation}
To optimize the model's performance and handle class imbalances, we applied several data augmentation techniques. After augmentation, we achieved desirable changes in the sample numbers for each class. The augmentation techniques we used include SMOTE, ADASYN, and custom augmentation.

Table ~\ref{tab:augmentation_results} presents the overall and per-class accuracy of the model trained on these augmented datasets. The best-performing augmentation technique, SMOTE, yielded the highest performance metrics, as highlighted in bold.

\subsection{Stage 3 Results: The Effect of Different Model Components}
Furthermore, we evaluated the impact of various components integrated into the final model architecture during Stage 3. The analysis was focused on the contributions of weight initialization strategies and the utilization of KAN in enhancing model performance.

\subsubsection{Effects of Weight Initialization}
Weight initialization plays a critical role in the training dynamics. We analyzed the consequences of omitting weight initialization for either the FIN or the N1 detector. The effects of these weight initialization strategies are depicted in Table ~\ref{tab:augmentation_results}.

\begin{figure}[ht]
    \centering
    \includegraphics[width=\linewidth]{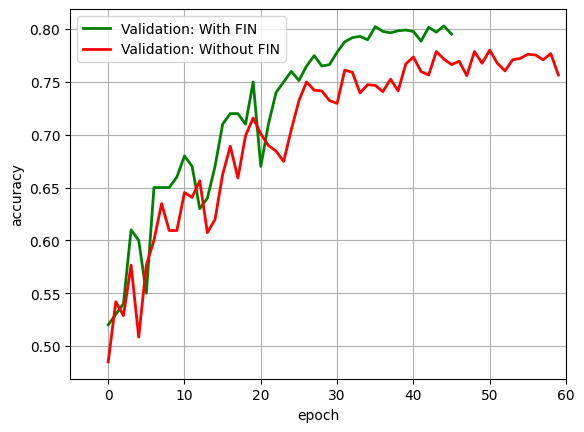}
    \caption{Validation Accuracy with FIN Initialization vs. without FIN Initialization.}
    \label{fig:compare}
\end{figure}

\begin{itemize}
    \item No Weight Initialization for FIN: Without the pretrained FIN, starting with random weights in Stage 3 resulted in a slower convergence rate as illustrated in Fig. ~\ref{fig:compare} and also a decrease in accuracy. This highlights the importance of pretraining in enhancing learning efficiency and model stability.
    \item No Weight Initialization for N1 Detector: Similarly, not initializing the N1 detector by omitting Stage 2 and beginning with random weights in Stage 3 led to slower convergence and lower accuracy, particularly for N1 identification. 
\end{itemize}

\begin{table}[H]
\centering
\caption{Classification Report for Final Model}
\begin{tabularx}{\linewidth}{X c c c c}
\toprule
\textbf{Class}           & \textbf{Precision} & \textbf{Recall} & \textbf{F1-Score} & \textbf{Support} \\
\midrule
Wake                     & 0.84               & 0.87            & 0.85              & 466              \\
N1                       & 0.72               & 0.60            & 0.66              & 275              \\
N2                       & 0.82               & 0.84            & 0.83              & 571              \\
N3                       & 0.77               & 0.85            & 0.81              & 99               \\
REM                      & 0.85               & 0.87            & 0.86              & 109              \\
\midrule
\textbf{Overall Accuracy} & -                  & -               & 0.81              & 1520             \\
\textbf{Macro Avg}        & 0.80               & 0.81            & 0.80              & 1520             \\
\textbf{Weighted Avg}     & 0.81               & 0.81            & 0.81              & 1520             \\
\bottomrule
\end{tabularx}
\label{tab:per_class_metrics}
\end{table}

\subsubsection{Comparison of KAN with a Multi-Layer Perceptron (MLP)}
The integration of KAN in the final architecture also had a significant impact on performance. When we replaced KAN with an MLP, the model's accuracy was lower (79.09\%). Notably, the inclusion of KAN added 119,424 parameters to the model, which likely contributed to its ability to capture complex relationships in the data more effectively than the traditional MLP. This highlights KAN’s superior capacity for modeling intricate patterns compared to MLPs.

\subsection{Stage 3 Results: Best Model Results}
We evaluated the final model on the non-augmented test dataset consisting of 3,000-sample ECG signals. The model trained on the SMOTE-augmented dataset achieved an overall best accuracy of 80.79\% and a Cohen's kappa score of 0.73. The per-class performance metrics are outlined in Table ~\ref{tab:per_class_metrics}.

\begin{figure}[ht]
    \centering
    \includegraphics[width=1\linewidth]{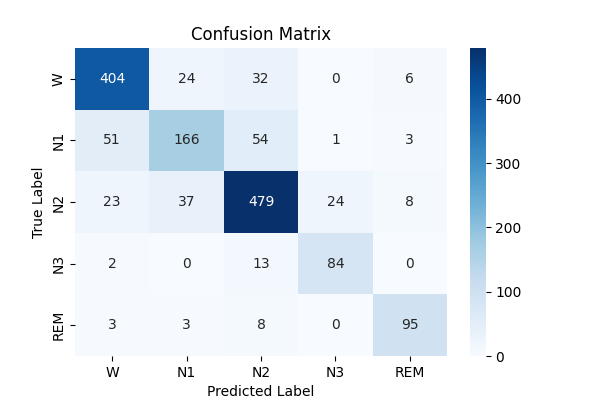}
    \caption{Confusion matrix of the sleep stage classification model on the test dataset. Each row represents the true label, and each column represents the predicted label. The diagonal values indicate correct classifications, while off-diagonal values represent misclassifications between sleep stages.}
    \label{fig:confusion_matrix}
\end{figure}

The confusion matrix in Fig. ~\ref{fig:confusion_matrix}, provides a detailed breakdown of how each sleep stage was classified by the model. This allows us to analyze the specific misclassifications between stages, particularly challenging transitions like N1 vs N2 and N1 vs REM. In Table \ref{tab:comparison_results}, we compare the performance of our proposed method with other ECG-based sleep stage classification studies.

\begin{table}[ht]
\centering
\caption{Comparison of Classification Accuracy in Previous Studies.}
\begin{tabularx}{\linewidth}{X c c c c c}
\toprule
\textbf{Reference}                    & \textbf{Accuracy} \\
\midrule
Lesmana et al. \cite{lesmana2023}     & 71.52\%           \\
Fikri et al. \cite{fikri2023}         & 73.38\%           \\
Matsumoto et al. \cite{matsumoto2020} & 74\%              \\
Surantha et al. \cite{surantha2023}   & 77\%              \\
Choi et al. \cite{choi2023}           & 76\%              \\
Pini et al. \cite{pini2022}           & 73\%              \\
Sridhar et al. \cite{sridhar2020}     & 80\%              \\
\midrule
\textbf{ECG-SleepNet}                 & 80.79\%           \\
\bottomrule
\end{tabularx}
\label{tab:comparison_results}
\end{table}

The model's complexity is reflected in its parameter count. The final integrated model, responsible for classifying all five sleep stages, has 4,140,421 parameters. Despite these substantial numbers, the model maintains efficient performance, with a latency of 97.77 milliseconds for processing a 30-second ECG signal, including preprocessing. This balance of accuracy and speed makes it suitable for real-time sleep stage classification applications.

\section{Discussions}
\label{sec:discussion}
The results of this study highlight several key contributions to improving sleep stage classification, particularly the role of FINs, the N1 detector, and data augmentation techniques. Together, these components significantly enhanced both the accuracy and efficiency of the model, addressing challenges such as class imbalances and the complexity of certain stages, especially N1.

The results from the FINs underscore their essential role in enhancing model performance. The removal of the FIN's initialized weights led to a decrease in classification accuracy. However, the most significant contribution of the FIN lies not only in improving accuracy but also in accelerating the convergence of the model toward its optimal performance. As illustrated in Figure ~\ref{fig:compare}, the model initialized with the pretrained FIN achieved its optimal performance approximately 10 epochs faster than its counterpart without FIN. Overall, these findings validated the utility of FINs in optimizing both the accuracy and efficiency of the training process in sleep stage classification tasks.

The introduction of the N1 detector in Stage 2 significantly enhanced the model's ability to classify the N1 stage, a notoriously challenging sleep phase due to its transitional nature. Removing this initialization led to a notable drop in N1 classification accuracy and overall performance, underscoring its critical role in capturing meaningful features specific to N1. This highlights the importance of specialized detection for accurately identifying difficult stages like N1, which can easily be confused with neighboring stages. In Stage 2, the N1 detector achieved an overall accuracy of 76.56\%, with a precision of 0.73 and a recall of 0.88 for the N1 class, as shown in ~\ref{tab:classification_report}. This combination of metrics highlights the N1 detector's critical role in improving classification performance, particularly for this transitional sleep stage, validating its integration into the overall architecture.

Sleep stage datasets are inherently imbalanced, which often leads to decreased model performance, particularly for classes with fewer samples. The application of data augmentation techniques, especially SMOTE, has demonstrated a positive impact on classification performance in our study. While the effectiveness of SMOTE oversampling for ECG signal augmentation remains unclear in general, our findings indicate that it significantly improved the model's ability to classify underrepresented stages like N1. This underscores the value of data augmentation in enhancing classification capabilities.

The model demonstrates strong classification performance across the majority of sleep stages. For instance, Wake (W) achieved a high accuracy of 86.7\%, with only minor misclassifications into adjacent stages, primarily N1 (5.1\%) and N2 (6.9\%). This highlights the model’s capacity to distinguish between wakefulness and sleep. Similarly, N2 and N3 stages were well classified, with accuracies of 83.9\% and 84.8\% respectively. For N3 (deep sleep), misclassifications primarily occurred into N2 (13.1\%), which is understandable given the physiological similarities between these stages. The REM stage also performed well, achieving 87.2\% accuracy, with most errors occurring in N2 (7.3\%), reinforcing the model’s robustness in distinguishing non-adjacent stages.

The N1 stage, being transitional, posed the greatest challenge with a 60.4\% accuracy, which aligns with the inherent difficulty of distinguishing this stage. Misclassifications into both Wake (18.5\%) and N2 (19.6\%) reflect the transitional nature of N1, as it is often confused with adjacent stages. However, this misclassification pattern indicates that the model is effectively capturing the natural progression between these stages rather than random errors, which is a positive indicator of robustness. Overall, the model performs well across the stages, particularly excelling in distinguishing between more distinct stages, with errors primarily confined to adjacent stages.

Clearly, the combination of FINs, the N1 detector, and data augmentation techniques significantly improved the performance of the sleep stage classification model. The results demonstrate not only enhanced classification accuracy but also faster convergence and greater stability in training. This suggests that the proposed architecture effectively addresses key challenges in sleep stage classification, particularly for stages that are inherently difficult to distinguish, leading to a more robust and reliable model for real-world applications.

Furthermore, it is important to note that the high classification performance was achieved using only single-channel ECG signals. This is particularly important given that most sleep stage classification studies rely on multi-channel EEG data. The ability to extract meaningful features and achieve robust classification with a minimal data input not only demonstrates the strength of the model architecture but also opens up possibilities for less invasive and more accessible sleep monitoring solutions.

\section{Conclusion}
\label{sec:conclusion}
In conclusion, we have presented a novel approach to sleep stage classification using single-channel ECG signals, achieving high performance without using traditional multi-channel EEG data. By integrating FINs and an N1-specific detector, alongside data augmentation techniques especially SMOTE, the model has addressed key challenges such as class imbalance and the accurate classification of transitional sleep stages. The enhanced convergence speed and classification accuracy highlight the efficiency and effectiveness of this architecture in tackling the complexities of sleep stage classification.

These results open the door for more practical and less intrusive sleep monitoring applications, with potential for wearable or home-based devices. The ability to achieve strong performance with minimal data input suggests a promising direction for future work, where expanding the model to incorporate additional physiological signals or applying it to broader populations could yield further opportunities in personalized sleep health solutions.

\bibliography{references}

\begin{thebibliography}{10}
\providecommand{\url}[1]{#1}
\csname url@samestyle\endcsname
\providecommand{\newblock}{\relax}
\providecommand{\bibinfo}[2]{#2}
\providecommand{\BIBentrySTDinterwordspacing}{\spaceskip=0pt\relax}
\providecommand{\BIBentryALTinterwordstretchfactor}{4}
\providecommand{\BIBentryALTinterwordspacing}{\spaceskip=\fontdimen2\font plus
\BIBentryALTinterwordstretchfactor\fontdimen3\font minus \fontdimen4\font\relax}
\providecommand{\BIBforeignlanguage}[2]{{%
\expandafter\ifx\csname l@#1\endcsname\relax
\typeout{** WARNING: IEEEtran.bst: No hyphenation pattern has been}%
\typeout{** loaded for the language `#1'. Using the pattern for}%
\typeout{** the default language instead.}%
\else
\language=\csname l@#1\endcsname
\fi
#2}}
\providecommand{\BIBdecl}{\relax}
\BIBdecl

\bibitem{blue_light}
M.~I. Silvani, R.~Werder, and C.~Perret, ``The influence of blue light on sleep, performance and wellbeing in young adults: A systematic review,'' \emph{Frontiers in Physiology}, vol.~13, p. 943108, 2022.

\bibitem{room_light}
J.~J. Gooley, K.~Chamberlain, K.~A. Smith, S.~B. Khalsa, S.~M. Rajaratnam, E.~Van~Reen, J.~M. Zeitzer, C.~A. Czeisler, and S.~W. Lockley, ``Exposure to room light before bedtime suppresses melatonin onset and shortens melatonin duration in humans,'' \emph{Journal of Clinical Endocrinology and Metabolism}, vol.~96, no.~3, pp. E463--E472, 2011.

\bibitem{rechtschaffen1968}
A.~Rechtschaffen and A.~Kales, \emph{A Manual of Standardized Terminology, Techniques and Scoring System for Sleep Stages of Human Subjects}.\hskip 1em plus 0.5em minus 0.4em\relax Brain Information Service/Brain Research Institute, 1968.

\bibitem{berry2012}
R.~B. Berry, R.~Brooks, R.~Cormack, A.~Culebras, C.~Dobson, E.~M., G.~M., D.~M., and J.~M., \emph{The AASM Manual for the Scoring of Sleep and Associated Events: Rules, Terminology, and Technical Specifications, Version 2.0}.\hskip 1em plus 0.5em minus 0.4em\relax American Academy of Sleep Medicine, 2012.

\bibitem{lyu2022}
J.~Lyu, W.~Shi, C.~Zhang, and C.~H. Yeh, ``A novel sleep staging method based on eeg and ecg multimodal features combination,'' \emph{IEEE Transactions on Neural Systems and Rehabilitation Engineering}, vol.~31, pp. 4073--4084, 2023, epub 2023 Oct 20.

\bibitem{tao2022}
S.~Tao, J.~Hu, W.~L. Goh, and Y.~Gao, ``Squeeze-excite fusion based multimodal neural network for sleep stage classification with flexible eeg/ecg signal acquisition circuit,'' in \emph{2024 IEEE International Symposium on Circuits and Systems (ISCAS)}, 2024.

\bibitem{utomo2023}
O.~K. Utomo, N.~Surantha, S.~M. Isa, and B.~Soewito, ``Automatic sleep stage classification using weighted elm and pso on imbalanced data from single lead ecg,'' \emph{Procedia Computer Science}, vol. 157, pp. 321--328, 2019.

\bibitem{lesmana2023}
T.~F. Lesmana, S.~M. Isa, and N.~Surantha, ``Sleep stage identification using the combination of elm and pso based on ecg signal and hrv,'' in \emph{2018 3rd International Conference on Computer and Communication Systems (ICCCS)}, 2018, pp. 258--262.

\bibitem{bozkurt2021}
F.~Bozkurt, M.~Uçar, C.~Bilgin \emph{et~al.}, ``Sleep–wake stage detection with single channel ecg and hybrid machine learning model in patients with obstructive sleep apnea,'' \emph{Phys Eng Sci Med}, vol.~44, pp. 63--77, 2021.

\bibitem{erdenebayar2020}
U.~Erdenebayar, Y.~Kim, J.-U. Park, S.~Lee, and K.-J. Lee, ``Automatic classification of sleep stage from an ecg signal using a gated-recurrent unit,'' \emph{International Journal of Fuzzy Logic and Intelligent Systems}, vol.~20, no.~3, pp. 181--187, 2020.

\bibitem{matsumoto2020}
H.~Matsumoto, S.~Okada, T.~Wang, and M.~Masaaki, ``Sleep stage estimation using ecg,'' in \emph{2020 IEEE International Conference on Bioinformatics and Biomedicine (BIBM)}.\hskip 1em plus 0.5em minus 0.4em\relax IEEE, 2020, pp. 2983--2983.

\bibitem{surantha2023}
N.~Surantha and V.~V. Jansen, \emph{Sleep-Stage Identification Using Recurrent Neural Network for ECG Wearable-Sensor System}.\hskip 1em plus 0.5em minus 0.4em\relax Auerbach Publications, 2023, pp. X--X.

\bibitem{fikri2023}
R.~Fikri and S.~M. Isa, ``Sleep stage classification on ecg signal and hrv using combination of random forest and shap value,'' \emph{INNOVATIVE: Journal Of Social Science Research}, vol.~3, no.~4, pp. 8680--8693, 2023.

\bibitem{pini2022}
N.~Pini, J.~L. Ong, G.~Yilmaz, N.~I. Y.~N. Chee \emph{et~al.}, ``An automated heart rate-based algorithm for sleep stage classification: validation using conventional psg and innovative wearable ecg device,'' \emph{medRxiv}, 2022.

\bibitem{choi2023}
I.~Choi and W.~Sung, ``Single-channel ecg-based sleep stage classification with end-to-end trainable deep neural networks,'' in \emph{2023 45th Annual International Conference of the IEEE Engineering in Medicine and Biology Society (EMBC)}, 2023.

\bibitem{mitbih}
Y.~Ichimaru and G.~B. Moody, ``Development of the polysomnographic database on cd-rom,'' \emph{Psychiatry and Clinical Neurosciences}, vol.~53, no.~2, pp. 175--177, April 1999.

\bibitem{goldberger2000}
A.~L. Goldberger, B.~E. Moody, R.~G. Mark, and et~al., ``Physiobank, physiotoolkit, and physionet: Components of a new research resource for complex physiologic signals,'' \emph{Circulation}, vol. 101, no.~23, pp. e215--e220, 2000.

\bibitem{DeChazal2003}
P.~De~Chazal \emph{et~al.}, ``Automated processing of the single-lead electrocardiogram for the detection of obstructive sleep apnoea,'' \emph{IEEE Transactions on Biomedical Engineering}, vol.~50, no.~6, pp. 686--696, 2003.

\bibitem{scipy2020}
P.~Virtanen \emph{et~al.}, ``{{SciPy} 1.0: Fundamental Algorithms for Scientific Computing in Python},'' \emph{Nature Methods}, vol.~17, pp. 261--272, 2020.

\bibitem{DBLP:journals/corr/abs-2110-04831}
S.~Saba-Sadiya, T.~Alhanai, and M.~M. Ghassemi, ``Feature imitating networks,'' in \emph{ICASSP 2022 - 2022 IEEE International Conference on Acoustics, Speech and Signal Processing (ICASSP)}, 2022, pp. 4128--4132.

\bibitem{SST}
D.-H. Pham and S.~Meignen, ``High-order synchrosqueezing transform for multicomponent signals analysis—with an application to gravitational-wave signal,'' \emph{IEEE Transactions on Signal Processing}, vol.~65, no.~12, pp. 3168--3178, 2017.

\bibitem{Muradeli2020ssqueezepy}
J.~Muradeli, ``ssqueezepy,'' GitHub repository, \url{https://github.com/OverLordGoldDragon/ssqueezepy/}, 2020.

\bibitem{resunet}
D.~Jha, P.~H. Smedsrud, M.~A. Riegler, D.~Johansen, T.~de~Lange, P.~Halvorsen, and H.~D. Johansen, ``Resunet++: An advanced architecture for medical image segmentation,'' 2019.

\bibitem{smote}
N.~V. Chawla, K.~W. Bowyer, L.~O. Hall, and W.~P. Kegelmeyer, ``Smote: Synthetic minority over-sampling technique,'' \emph{Journal of Artificial Intelligence Research}, vol.~16, p. 321–357, 2002.

\bibitem{adasyn}
H.~He, Y.~Bai, E.~A. Garcia, and S.~Li, ``Adasyn: Adaptive synthetic sampling approach for imbalanced learning,'' in \emph{2008 IEEE International Joint Conference on Neural Networks (IEEE World Congress on Computational Intelligence)}, 2008, pp. 1322--1328.

\bibitem{kolmogorov1957}
A.~N. Kolmogorov, ``On the representation of continuous functions of many variables by superposition of continuous functions of one variable and addition,'' \emph{Doklady Akademii Nauk}, vol. 114, pp. 953--956, 1957.

\bibitem{liu2024kankolmogorovarnoldnetworks}
Z.~Liu, Y.~Wang, S.~Vaidya, F.~Ruehle, J.~Halverson, M.~Soljačić, T.~Y. Hou, and M.~Tegmark, ``Kan: Kolmogorov-arnold networks,'' 2024.

\bibitem{sridhar2020}
N.~Sridhar, A.~Shoeb, P.~Stephens \emph{et~al.}, ``Deep learning for automated sleep staging using instantaneous heart rate,'' \emph{npj Digital Medicine}, vol.~3, no.~1, p. 106, 2020.

\end{thebibliography}

\end{document}